\documentclass[preprint,12pt]{elsarticle}

\usepackage{graphicx}
\usepackage{latexsym}
\usepackage{times}
\usepackage{soul}
\usepackage{url}
\usepackage[hidelinks]{hyperref}
\usepackage[utf8]{inputenc}
\usepackage[small]{caption}
\usepackage{graphicx}
\usepackage{amsmath}
\usepackage{amssymb}
\usepackage{booktabs}
\usepackage{graphicx}
\usepackage{graphics}
\usepackage{multirow}
\usepackage[dvipsnames]{xcolor}
\usepackage{subfig}
\usepackage{algorithm}
\usepackage{listings}
\journal{Knowledge-Based Systems}
\usepackage{hyperref}
\usepackage{lineno}

\begin{document}

\begin{frontmatter}

\title{Cross-Domain Few-Shot Learning via Adaptive Transformer Networks}

\author[Naeem]{Naeem Paeedeh \fnref{fn1}}
\author[Dhika]{Mahardhika Pratama\fnref{fn1}}
\author[Anwar]{Muhammad Anwar Ma'sum}
\author[Wolfgang]{Wolfgang Mayer}
\author[Jimmy]{Zehong Cao}
\author[Ryszard,Ryszard2]{Ryszard Kowlczyk}

\fntext[fn1]{Authors share equal contributions.}

\address[Naeem]{STEM, University of South Australia, Adelaide, Australia. e-mail: naeem.paeedeh@mymail.unisa.edu.au}
\address[Dhika]{STEM, University of South Australia, Adelaide. Australia. e-mail: dhika.pratama@unisa.edu.au}
\address[Anwar]{STEM, University of South Australia, Adelaide. Australia. e-mail: muhammad\_anwar.masum@mymail.unisa.edu.au}
\address[Wolfgang]{STEM, University of South Australia, Adelaide. Australia. e-mail: Wolfgang.Mayer@unisa.edu.au}
\address[Jimmy]{STEM, University of South Australia, Adelaide. Australia. e-mail: Jimmy.Cao@unisa.edu.au}
\address[Ryszard]{STEM, University of South Australia, Adelaide. Australia. e-mail: Ryszard.Kowalczyk@unisa.edu.au}
\address[Ryszard2]{Systems Research Institute, Polish Academy of Sciences, Poland}

\begin{abstract}
Most few-shot learning works rely on the same domain assumption between the base and the target tasks, hindering their practical applications. This paper proposes an adaptive transformer network (ADAPTER), a simple but effective solution for cross-domain few-shot learning where there exist large domain shifts between the base task and the target task. ADAPTER is built upon the idea of bidirectional cross-attention to learn transferable features between the two domains. The proposed architecture is trained with DINO to produce diverse, and less biased features to avoid the supervision collapse problem. Furthermore, the label smoothing approach is proposed to improve the consistency and reliability of the predictions by also considering the predicted labels of the close samples in the embedding space. The performance of ADAPTER is rigorously evaluated in the BSCD-FSL benchmarks in which it outperforms prior arts with significant margins.

\end{abstract}


\end{frontmatter}


\section{Introduction}
The advent of deep learning (DL) technologies has revolutionized the field of Artificial Intelligence (AI), making possible for complex data to be processed in an end-to-end manner without a manual feature extraction step. Although DL technologies have delivered state-of-the-art performances in various application domains, their successes are mainly attributed to large sample sizes hindering their applications in resource-constrained environments and being hardly applicable in the low sample setting.

The few-shot learning problem is motivated by the issue of data scarcity, where the main goal is to adapt quickly to new classes with extremely few samples. Despite its progress, most few-shot learning solutions assume that the base task and the target task are drawn from the same domain and often fail when there exist extreme discrepancies between the base task and the target task~\cite{Guo2019ANB,Phoo2021SelftrainingFF}, i.e., they are inferior to naive fine-tuning strategies in the cross-domain cases. Extreme discrepancies here refer to different image types, such as ImageNet as the base task and satellite or X-ray images as the target task~\cite{Phoo2021SelftrainingFF}. 

We follow the same setting as~\cite{Guo2019ANB,Islam2021DynamicDN,Phoo2021SelftrainingFF}, where such a problem is overcome with moderate amounts of unlabelled samples from the target domain. The existence of unlabelled samples makes it possible to apply self-supervised learning techniques to avoid the supervision collapse problem \cite{Doersch2020CrossTransformersSF}, where it loses necessary information for knowledge transfer. That is, the self-supervised learning step is capable of extracting meaningful features for generalizations beyond the base classes. The self-supervised learning strategy alone is not sufficient because it calls for a large number of unlabelled samples. Although unlabelled samples are given, the cross-domain few-shot learning problems distinguish themselves from the classical domain adaptation problem \cite{Pan2010ASO} because the base and the target task possess disjoint classes. The combination of labelled samples of the base task and unlabelled samples of the target task is a feasible solution because the labelled samples allow for the construction of common features while generalization beyond the base classes is availed by the unlabelled samples \cite{Islam2021DynamicDN}. 

To the best of our knowledge, there exist several prior works in this domain where \cite{Phoo2021SelftrainingFF} adopts the concept of self-training for unlabelled samples and SimCLR \cite{Chen2020ASF} for representation learning while \cite{Islam2021DynamicDN} adopts the concept of consistency regularization for weakly augmented unlabelled images and strongly augmented unlabelled images. The task adversarial augmentation method is proposed in \cite{Wang2021CrossDomainFC}, where the adversarial learning concept is implemented to construct the virtual tasks presenting the worst-case problem. \cite{Tseng2020CrossDomainFC} proposes the feature-wise transformation layers simulating different feature distributions across different domains. \textcolor{black}{Knowledge Transduction (KT) is proposed in \cite{Li2023KnowledgeTF} where it relies on the feature adaptation module and the stochastic image transformation. \cite{Zhao2023DualAR} proposes the dual adaptive representation alignment approach and \cite{Fu2023StyleAdvMS} addresses the cross-domain few-shot learning problem with the style adversarial training method.} \textcolor{black}{This problem remains an open problem because these works do not yet explore any domain alignment steps. The generalization power of the target task is upper bounded by the distance of the base task and the target task~\cite{BenDavid2006AnalysisOR} in addition to the model's performances in recognizing the base task, which holds even in the case of mutually exclusive tasks~\cite{Fang2021OpenSD}. A recent finding in \cite{Hu2022PushingTL} demonstrates significant performance gains with the pre-training step of the base task. Besides, all algorithms are crafted from the convolutional neural network's (CNN) solution, performing sub-optimally compared to that of the transformer approaches. In summary, two research gaps are to be attacked in this paper: 1) the issue of domain shifts remains an open problem in the literature; 2) existing approaches are still crafted under the convolutional neural network backbone.}   

This paper proposes an adaptive transformer network (ADAPTER) for cross-domain few-shot learning problems. It is built upon a compact transformer (CCT) backbone~\cite{Hassani2021EscapingTB}. The CCT is lighter than the ViT \cite{Dosovitskiy2021AnII} because of the convolutional tokenizer. Our model is structured under the bidirectional cross-attention transformer~\cite{Wang2022DomainAV}, making use of the cross-attention layer \cite{Xu2022CDTransCT} to match the target queries and the base-task keys for smooth knowledge transfers. This strategy induces blurry task boundaries, facilitating seamless knowledge transfers. Our experiments follow the BSCD-FSL benchmark protocol \cite{Guo2019ANB} where the target task not only features different domains of the base task but also presents different image types, e.g., internet images vs X-ray images. To make this problem feasible, a model gains access to unlabelled samples of the target task.  

ADAPTER is simple but highly effective, where the first phase is to perform the self-supervised training step of the base task via the bidirectional transformer structure using the DINO strategy \cite{Caron2021EmergingPI}. The DINO strategy is selected because it is designed specifically for the transformer backbone.
The self-supervised learning strategy extracts general features to prevent the supervision collapse problem \cite{Hu2022PushingTL,Doersch2020CrossTransformersSF}, while bidirectional features function as implicit domain alignment steps to dampen the gaps of the base task and the target task \cite{Wang2022DomainAV}. This structure also enables both base samples and target samples to be utilized simultaneously under a single training stage, thus reducing the number of training stages. We do not apply any supervised training step because such an approach results in the overfitting problem of the base classes \cite{Lu2022SelfSupervisionCB}. 

The self-supervised learning step using DINO under the bidirectional transformer architecture is followed by the supervised learning step of the support set comprising very few labelled samples to create a new classification head. Since a low sample quantity causes biased classifier responses, the label smoothing mechanism is undertaken to correct the classifier's logits, thus improving generalizations. The label smoothing mechanism is developed from the notion of label propagation \cite{Zhou2003LearningWL} trading off spatial locations of an image in the embedding space and original predictions. It combines both distance measures and the classifier's logits in inferring the class labels. Unlike \cite{Phoo2021SelftrainingFF}, no self-training mechanism is applied in our framework because the projection of the target task to the base label space leads to sub-optimal solutions. 

\noindent This paper puts forward at least four major contributions:

\begin{enumerate}
    \item ADAPTER is proposed in this paper as a simple but effective approach to address the cross-domain few-shot learning problems.
    \item This paper proposes the self-supervised learning step via DINO \cite{Caron2021EmergingPI} under the bidirectional transformer architecture, making it possible for base features and target features to be exploited simultaneously.
    \item The label smoothing mechanism is developed to address biased responses of classifiers due to low sample sizes of the support set.
    \item Source codes of ADAPTER are provided \footnote{\href{https://github.com/Naeem-Paeedeh/ADAPTER}{https://github.com/Naeem-Paeedeh/ADAPTER}} to assure reproducibility and convenient further study.
\end{enumerate}

\textcolor{black}{Our novelty is present in two aspects: 1) the proposed approach applies the bidirectional transformer architecture built upon the quadruple transformer block and the compact transformer (CCT) backbone for the cross-domain few-shot learning problems; 2) the self-supervised learning strategy via DINO is implemented under the bidirectional transformer architecture allowing the base task and the target task to be exploited simultaneously. Such mechanism functions as a domain alignment step minimizing the distance of the base task and the target task. }


\textcolor{black}{The advantage of ADAPTER lies in its simplicity for cross-domain few-shot learning problems where such problem is handled with the self-supervised learning phase via DINO under the bidirectional transformer architecture coupled with the label smoothing strategy. Empirically, the advantage of ADAPTER is also evident where our approach outperforms prior arts in the BSCD-FSL benchmark with $1-9\%$ gaps in 14 out of 16 experiments. }

\section{Related Works}
\noindent\textbf{Few Shot Learning} is a learning problem to address the issue of data scarcity in dynamic environments, making it possible for a model to adapt quickly to new environments with extremely few samples. Existing works are categorized into two groups: metric-based and adaptation-based approaches. MatchingNet \cite{Vinyals2016MatchingNF} is perceived as a pioneering metric-based approach where cosine similarity measures outputs classification outcomes. The relation network \cite{Sung2018LearningTC} is another early example of the metric-based approach, learning its own similarity metric. Prototypical Networks (ProtoNet) \cite{Snell2017PrototypicalNF} is a popular few-shot learning approach making use of the prototype concept of the support set to produce the classification decision. On the other hand, model-agnostic meta-learning \cite{Finn2017ModelAgnosticMF} is a prominent adaptation-based approach finding an initialization strategy through the meta-learning approach, enabling fast adaptations to new concepts with few gradient steps. \cite{Sun2019MetaTransferLF} exemplifies the adaptation-based approach for few-shot learning benefiting from the meta-learning concept and the parameter transformation strategy. All these works have been successful when the base task and the target task are from the same domain but fail in the case of cross-domain few-shot learning \cite{Phoo2021SelftrainingFF}. 

BSCD-FSL benchmark \cite{Guo2019ANB} has been proposed as a challenging problem of few-shot learning where large domain gaps exist between the base and the target task. They also demonstrate the failure cases of popular few-shot learning approaches performing even worse than the naive fine-tuning strategy. \cite{Phoo2021SelftrainingFF} presents the STARTUP method derived from the combination of self-training and self-supervised learning techniques. \cite{Islam2021DynamicDN} enhances this method with the weak and strong augmentation concept claimed as a self-supervised learning approach and removes the self-training step. \cite{Wang2021CrossDomainFC} puts forward the task adversarial augmentation method to create the virtual tasks featuring the worst-case problem, and \cite{Tseng2020CrossDomainFC} proposes the feature-wise transformation layers simulating various feature distributions under different domains. \textcolor{black}{In \cite{Li2023KnowledgeTF}, the knowledge transduction approach is proposed while \cite{Zhao2023DualAR} presents the dual adaptive representation alignment approach. \cite{Fu2023StyleAdvMS} puts forward the style adversarial training approach for cross-domain few-shot learning problem.} These approaches still neglect the domain alignment step, which might boost the performance even further. Besides, these methods are still based on the CNN structure, which might be sub-optimal compared to the transformer solution. The transformer solutions for few-shot learning have appeared in \cite{Doersch2020CrossTransformersSF,Liu2021AUR}. We argue that these methods do not yet utilize the full potential of the ViT architecture \cite{Dosovitskiy2021AnII}. In addition, they are tested with the meta-dataset \cite{Triantafillou2020MetaDatasetAD} rather than the BSCD-FSL dataset. \textcolor{black}{This paper aims to cope with two research issues: 1) the domain alignment step remains an open research problem in the cross-domain few-shot learning problem; 2) existing approaches are still built upon the convolutional neural network structure.}

\noindent\textbf{Unsupervised Domain Adaptation} is closely related to our work where the goal is to adapt a model to a new domain without any supervision of the target domain \cite{BenDavid2006AnalysisOR,Pan2010ASO,Ganin2016DomainAdversarialTO,Fang2021OpenSD}. Recently, transformer-based approaches have been proposed for unsupervised domain adaptation \cite{Wang2022DomainAV,Xu2022CDTransCT,Yang2021TVTTV}. Our work differs from them because the base task and the target task present distinct target classes. To the best of our knowledge, we are the first to explore the application of domain alignment for cross-domain few-shot learning under the transformer backbone, taking advantage of the presence of moderate unlabelled samples of the target task in addition to the self-supervised strategy.

\section{Problem Formulation}
The cross-domain few-shot learning problem aims to classify unknown classes of the target task $\mathcal{D}_{T}$ over $\mathcal{X}_{T}\times\mathcal{Y}_{T}$ based on very few labelled samples $\mathcal{S}=\{x_i,y_i\}_{i=1}^{N_S} \backsim \mathcal{D}_{T}$ known as the support set where $x_i\in\mathcal{X}_{T}$ and $y_i\in\mathcal{Y}_{T}$ respectively denote an $i-th$ input image and its corresponding label. An evaluation is performed in the unlabelled query set $Q=\{x_i\}_{i=1}^{N_Q}\backsim\mathcal{D}_{T}$ drawn from the same domain as the support set. Before the few-shot learning phase begins, one is given access to a base task $\mathcal{X}_{B}\times\mathcal{Y}_{B}\in\mathcal{D}_{B}$ where there exist plenty labelled samples $B=\{x_i,y_i\}_{i=1}^{N_B}\backsim\mathcal{D}_{B}$. The unique aspect of this problem lies in the existence of an extreme gap between the base task and the target task, i.e., completely distinct image space and label space. To make this problem feasible, unlabelled samples of the target task are supplied $U=\{x_i\}_{i=1}^{N_u}\backsim\mathcal{D}_{T}$ independent from the support set $S$ and the query set $Q$. In a nutshell, a model is developed through two learning phases: representation learning via $B,U$ and few-shot learning via $S$. The final evaluation is done with $Q$.

\section{ADAPTER: Network Structure}
\begin{figure}[!t]
 \centering
 \includegraphics[scale=0.55]{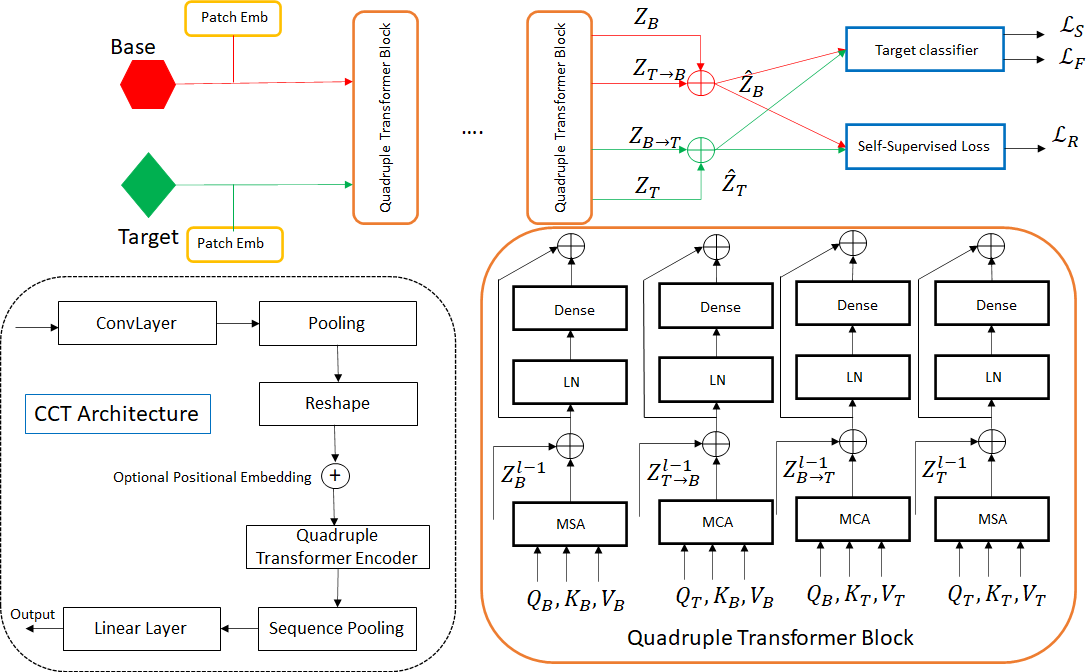}
 \caption{ADAPTER learns the base task $B$ and the unlabelled samples of the target task $U$ in a self-supervised manner using the DINO method and bidirectional features of the quadruple transformer block. A target classification head is created using a few labelled samples of the target task $S$. The label smoothing procedure via the label propagation method is carried out to refine the model's predictions.}
 \label{fig_framework}
 \end{figure}

ADAPTER is developed with the bidirectional cross-attention transformer \cite{Wang2022DomainAV} modifying the CCT backbone \cite{Hassani2021EscapingTB} with bidirectional cross-attention layers \cite{Xu2022CDTransCT}. The CCT backbone differs from the original ViT \cite{Dosovitskiy2021AnII} with the presence of a convolutional tokenizer comprising convolutional and pooling layers downsizing the patch size thus improving scalability. The network structure of ADAPTER putting forward the quadruple transformer block is portrayed in Fig. \ref{fig_framework}. That is, it relies on the self-attention layer processing the single domain features, i.e., base and target and the cross-attention layer mining the cross-domain features, i.e., base$\rightarrow$target and target$\rightarrow$base. Both of them capture long-ranges dependencies of features of interests. 

As with the conventional ViT, an input image $x_i\in\Re^{H\times W\times C}$ is converted into a sequence of image patches $\hat{x}_i\in\Re^{N\times(P^2C)}$ where $H,W,C$ are respectively the heights, weights and number of channels of the original images while $(P,P)$ is the resolution of image patches and $N=\frac{HW}{P^2}$ is the number of image patches. $\hat{x}_i$ is transformed respectively into queries, keys and values $Q\in\Re^{N\times d_k}, K\in\Re^{N\times d_k}, V\in\Re^{N\times d_v}$ using linear projectors $\Psi_Q\in\Re^{d_k},\Psi_K\in\Re^{d_k},\Psi_V\in\Re^{d_v}$ where $d_k,d_v$ stand for the dimensions of linear projectors. The self-attention layer operates in a single domain receiving $x_i^{s}$ of the base task $\mathcal{D}_{B}$ or $x_i^{T}$ of the target task $\mathcal{D}_{T}$.
\begin{equation}
    Attn_{B}(x_i^{B})=softmax(\frac{Q_{B}K_{B}^{'}}{\sqrt{d}_k})V_{B}
\end{equation}
\begin{equation}
    Attn_{T}(x_i^{T})=softmax(\frac{Q_{T}K_{T}^{'}}{\sqrt{d}_k})V_{T}
\end{equation}
where $^{'}$ denotes the transpose operation. The self-attention layer performs a weighted sum of the values where its weights offer the compatibility degrees between the queries and the keys. In other words, relevant input patches can be favoured over others.  
The cross-attention layer distinguishes itself from the self-attention layer where it explores matching degrees of base task and target task rather than self similarities. That is, the queries $Q\in\Re^{M\times d_k}$ are obtained from different tasks of those the keys $K\in\Re^{N\times d_k}$ and the values $V\in\Re^{N\times d_v}$. This strategy eases the alignment of the two tasks because similar patches across the two tasks should contribute more to the outputs than those of dissimilar input patches.  
\begin{equation}
    Attn_{T\rightarrow B}(x_i^{T},x_i^{B})=softmax(\frac{Q_{T}K_{B}^{'}}{\sqrt{d}_k})V_{B}
\end{equation}
\begin{equation}
    Attn_{B\rightarrow T}(x_i^{B},x_i^{T})=softmax(\frac{Q_{B}K_{T}^{'}}{\sqrt{d}_k})V_{T}
\end{equation}
The weights assigned to the values are higher in the case of similar samples than those of dissimilar samples \cite{Xu2022CDTransCT}. The cross-attention layer generates mixed features blurring the boundaries of the tasks, thus aligning their features \cite{Wang2022DomainAV}. 

The self-attention layer or the cross-attention layer is followed by a normalization layer, and a stack of dense layers parameterized with $\Psi_{W}$ using the GeLU activation functions. In addition, two residual connections are integrated here to prevent the vanishing/exploding gradient problems where the first part combines the outputs of self-attention or cross-attention layers with the outputs of a previous quadruple transformer block while the second part adds the outputs of the Multi-Layer Perceptron (MLP) layer with the outputs of the self-attention or cross-attention layer as per the original ViT. We follow the original CCT backbone \cite{Hassani2021EscapingTB} with a stack of quadruple transformer blocks forming a transformer encoder whose output is aggregated to the sequence pooling operator before going to a linear layer for predictions. 
\section{ADAPTER: Learning Policy}
ADAPTER consists of two learning phases: the representation learning phase and the few-shot learning phase where the representation learning phase performs the domain alignment step via the self-supervised training strategy of bidirectional features to generalize beyond known base classes using $B$ and $U$ while the few-shot learning phase adapts the label space of a model using extremely few labelled samples of $S$. Our model is structured as the bidirectional cross-attention transformer where $\zeta_{\Psi}(.)$ stands for the feature extractor consisting of a stack of the quadruple transformer blocks and $\xi_{\Phi}(.)$ denotes a classifier. $\Psi=\{\Psi_{Q},\Psi_{K},\Psi_{V},\Psi_{W}\}$ stand for the parameters of the feature extractor and $\Phi$ is the parameters of the classifier. The feature extractor generates the base features $Z_{B}=\zeta_{\Psi}(x_{i}^{B})$, the target-base features $Z_{T\rightarrow B}=\zeta_{\Psi}(x_{i}^{T},x_{i}^{B})$, the base-target features $Z_{B\rightarrow T}=\zeta_{\Psi}(x_{i}^{B},x_{i}^{T})$ and the target features $Z_{T}=\zeta_{\Psi}(x_{i}^{T})$ while the classifier converts the augmented feature vector to the label space of the target task $\xi_{\Phi}(\hat{Z})\in\Re^{D}\rightarrow\Delta_{M_{T}}$. $M_T, D$ denote the dimension of target classes and embedding space.
\subsection{Representation Learning Phase}
The representation learning phase starts with the self-supervised learning step of the base and target tasks simultaneously $B\backsim\mathcal{D}_B,U\backsim\mathcal{D}_T$ using the quadruple transformer block as an implicit domain adaptation step. This strategy follows the finding of \cite{Hu2022PushingTL} where the pre-training phase with an external dataset significantly boosts the performances of few-shot learners and differs from existing cross-domain few-shot learning methods \cite{Phoo2021SelftrainingFF,Islam2021DynamicDN} where such approaches are applied to the target domain. This is applied under the bidirectional architecture utilizing both base and target features, thus functioning as an implicit domain adaptation phase. In addition, no supervised step is applied on the base task because it leads to the overfitting problem of base classes \cite{Lu2022SelfSupervisionCB}. The self-supervised learning step produces general features which can be easily adapted to another domain and prevents the supervision collapse problem. We leverage the DINO method here \cite{Caron2021EmergingPI} because it is designed specifically for the transformer backbone. 

The DINO method utilizes the knowledge distillation approach \cite{Hinton2015DistillingTK} between the teacher network $\xi_{\Phi_{t}}(.)$ and the student network $\xi_{\Phi_{s}}(.)$ using the multi-crop strategy resulting in different distorted views of an image $V$. That is, two global views of an image, $x_1^{g}$ and $x_2^{g}$, are fed to the teacher only while all crops are passed to the student network. This is done with augmented feature space via the quadruple transformer block $\hat{Z}=[Z_{B},Z_{T\rightarrow B},Z_{B\rightarrow T},Z_{T}]$ where $Z_{T\rightarrow B}$ and $Z_{B\rightarrow T}$ stand for base-dominant features and target-dominant features respectively. The target features are aligned to the base where the target query $Q_{T}$ is matched with the base keys $K_{B}$ and values $V_{B}$ for base-dominant features and vice versa for target-dominant features. The loss function is expressed as:
\begin{equation}\label{base}
\begin{split}
    \mathcal{L}_{R}=\mathbb{E}_{(x_i^{B})\backsim\mathcal{D}_{B}\cup(x_i^{T})\backsim\mathcal{D}_{T}}[\\\sum_{x\in\{x_{1}^{g},x_{2}^{g}\}}\sum_{\tilde{x}\in V}l(\xi_{\Phi_{t}}(\zeta_{\Psi_{t}}(x_i^{T},x_i^{B}),\xi_{\Phi_{s}}(\zeta_{\Psi_{s}}(\tilde{x}_i^{T},\tilde{x}_i^{B}))]
\end{split}    
\end{equation}
where $l(.)$ denotes the cross-entropy loss function and $x\neq\tilde{x}$. Note that different temperatures, $\tau_s,\tau_t$, are used in the cross entropy loss function. The same architecture is implemented for both students and teachers. \eqref{base} is minimized to adjust the student network, i.e., the stop gradient technique is applied to the teacher. The momentum encoder technique following the exponential moving average update rule is adopted to adjust the teacher network. Let $\theta=[\Phi,\Psi]$, $\theta_t=\lambda\theta_t+(1-\lambda)\theta_s$ where $\lambda$ is dynamically adjusted with the cosine scheduler. As with DINO \cite{Caron2021EmergingPI}, the teacher outputs are centered and sharpened to avoid model collapse. The teacher parameters are copied and passed to the few-shot learning phase.  

\subsection{Few-Shot Learning Phase}
The few-shot learning phase is carried out by mining few labelled samples of the target domain $S$ to construct a new classification head. First, the new classification head is adopted for the classifier $\xi_{\Phi}(.):\mathcal{X}_{T}\rightarrow\Re^{M_T}$ to adapt to the label space $\mathcal{Y}_{T}$. Scarcely labelled samples $S$ are learned in the supervised manner via the cross-entropy loss function.
\begin{equation}
\begin{split}
    \mathcal{L}_{S}=\mathbb{E}_{(x_i^{T},y_i^T)\backsim\mathcal{D}_{T}}[l(\xi_{\Phi}(\zeta_{\Psi}(x_i^{T}),y_i^{T}))]
\end{split}
\end{equation}
where only self-attention features are used due to the sample size constraint. Since the base task and the target task share no overlapping label spaces $\mathcal{Y}_{B}\cap\mathcal{Y}_{T}=\emptyset$, the pseudo-labelling strategy w.r.t the projection of the base task as done in \cite{Phoo2021SelftrainingFF} might be sub-optimal. Because of a tiny sample size in $S$, this step only does not suffice and often suffers from biased classifier responses. 


 The label smoothing step is implemented here by applying the label propagation strategy \cite{Zhou2003LearningWL} enforcing the consistency of the predictions. That is, neighboring samples must be labelled similarly or be smooth. To this end, a graph $G=(V,E)$ is created where the nodes $V$ are defined by all images in $Q$ and $E$ stands for the edges weighted by an adjacency matrix $A\in\Re^{N_Q\times N_Q}$ whose elements are calculated as a proximity of two images in the embedding space $a_{i,j}=\exp{(\frac{-||\zeta_{\Psi}(x_i^{U},x_i^{U})-\zeta_{\Psi}(x_j^{U},x_j^{U})||^2)}{2\sigma^{2}})}, j\neq i$ and $a_{i,i}=0$. $\sigma$ denotes the standard deviation of the embedded images. The adjacency matrix is normalized as per $\hat{A}=D^{-1/2}AD^{-1/2}$ in which $D$ is a diagonal matrix whose the $(i,i)-th$ element denotes the sum of the $i-th$ row of $A$. The label propagation is iterated until convergence $\mathcal{F}^*$. 

\begin{equation}\label{label_propagation}
    \mathcal{F}_{t+1}=\alpha\hat{A}\mathcal{F}_{t}+(1-\alpha)\overline{Y}
\end{equation}
where $\mathcal{F}\in\Re^{N_Q\times M_T}$ denotes the label assignment matrix mapping the input space to the label space $\mathcal{F}:\mathcal{X}_{T}\rightarrow\Re^{M_T}$ and $\overline{Y}\in\Re^{N_Q\times M_T}$ stands for the initial label assignment matrix set as those of pseudo labels. The elements of $\overline{Y}$ is set as a one-hot encoded form of a pseudo label $\overline{y}_{i}$. $\alpha\in [0,1]$ is a hyper-parameter controlling a trade-off between the smoothness constraint, i.e., a desirable classification decision should not deviate too much across neighboring points, and the fitting constraint, i.e., a desirable classification decision should not differ from its initial label assignment \cite{Zhou2003LearningWL}. A closed-loop solution can be derived for \eqref{label_propagation}:
\begin{equation}
    \mathcal{F}^*=(I-\alpha\hat{A})^{\dagger}\overline{Y}
\end{equation}
where $^{\dagger}$ is the pseudo-inverse symbol. A final prediction is taken from the label assignment matrix $\hat{y}_{i}=\arg\max_{m\in M_{B}}\mathcal{F}_{i}^{*}, \{x_i,\overline{y}_i\}\in P$. Note that no label of evaluation samples in $Q$ is leaked for predictions. For the testing phase on $Q$, ADAPTER operates in the target task without any base samples. The double target strategy \cite{Wang2022DomainAV} is implemented where the cross-attention layer executes target features rather than cross features, thus leading to $\hat{Z}=[Z_{T},Z_{T},Z_{T},Z_{T}]$. Pseudo-code of ADAPTER is presented in Algorithm \ref{algo:ADAPTER}. 
\begin{algorithm}[tb]
		\caption{PyTorch style pseudocode for ADAPTER}
		\label{algo:ADAPTER}
		\definecolor{codeblue}{rgb}{0.25,0.5,0.5}
		\lstset{
			basicstyle=\fontsize{7.2pt}{7.2pt}\ttfamily\bfseries,
			commentstyle=\fontsize{7.2pt}{7.2pt}\color{codeblue},
			keywordstyle=\fontsize{7.2pt}{7.2pt},
		}
		\begin{lstlisting}[language=python]
def representation_learning(student):
    teacher = deepcopy(student)
    # iter_target: An iterator that loads the next mini-batch from the target dataset
    optimizer = torch.optim.AdamW(studend.parameters())
    
    for i in range(epochs):
        for xb in data_loader_base:
            xt = next(iter_target)
            # yt and ys prefixes are the outputs of the teacher and students. 
            # b, tb, t, and bt suffixes denote the base, target-base, target, and base-target features
            # Inputs are the 2 global views    
            yt_b, yt_tb, yt_t, yt_bt = teacher(xb[:2], xt[:2])
            ys_b, ys_tb, ys_t, ys_bt = student(xb, xt)
            loss_base = dino_loss_base(cat(ys_b, ys_tb), cat(yt_b, yt_tb))
            loss_target = dino_loss_target(cat(ys_t, ys_bt), cat(yt_t, yt_bt))
            loss = loss_base + loss_target
            loss.backward()
            optimizer.step()
            # Updates the teacher's parameters with an exponential moving average of momentum, centering, 
            # and sharpening
            update_teacher(teacher)
    teacher_cct = unwrap(teacher)   # Extracts only the CCT backbone
    return teacher_cct
            
def evaluation(cct):
   for support_set, query_set in episodic_data_loader:
        cct.reset()
        classifier = MLP(4*embed_dim, n_hidden, n_ways)
        optimizer = SGD(classifier.params())
        for i in range(n_epochs):
            for xs, ls in support_set:  # for each sample and label mini-batch
                features_s = cct(xs)
                ys = classifier(cat(features_s, features_s, features_s, features_s))
                loss = cross_entropy_loss(ys, ls)
                loss.backward()
                optimizer.step()
        xq, lq = query_set  # Images and labels of the query set
        features_q = cct(xq)
        logits_q = classifier(features_q)
        smoothed_logits = LabelPropagation(features_q, logits_q)
        list_correct.append(correct_predictions(smoothed_logits, lq))
    calculate_overall_accuracy(list_correct)

def ADAPTER():
    cct = CCT_14_7x2_with_quadruple_blocks() # Initializing a CCT-14/7x2 with quadruple blocks
    # head_base: MLP head for the base domain
    # head_target: MLP head for the target domain
    # Wrapping the cct, head_base, and head_target wrapped in one network object
    student = wrap(cct, head_base, head target)
    teacher = representation_learning(student)
    evaluation(teacher)
        \end{lstlisting}
    \end{algorithm}

\section{Time Complexity Analysis}
We utilize a CCT with quadruple blocks in our experiments. The most time-consuming parts of our model are self-attentions, cross-attentions, MLP projections in the blocks, sequence pooling, and MLP head. In the following, we calculate the time complexity for each component. At last, we can find the total time complexity for the CCT.\par
First, we have four combinations for self-attention and cross-attention. Each self-attention or cross-attention requires a linear projection to calculate query, key, and value. The final attention calculations need three consecutive matrix multiplications. Therefore, the time complexity for attentions is $O(nd^2)$, where $n$ is the number tokens or sequence length and $d$ is the dimension of each token or embedding. Second, for each MLP that has $L_{\text{MLP}}$ layers and $d_{\text{MLP}}$ neurons in hidden layers, we approximately have $L_{\text{MLP}}$ matrix multiplications. The time complexity for MLP is $O(L_{\text{MLP}}d_{\text{MLP}}^2)$. For $L_{\text{CCT}}$ layers or blocks, the time complexity of MLP is $O(L_{\text{CCT}}L_{\text{MLP}}d_{\text{MLP}}^2)$. We have a few convolution layers for patch embedding. According to \cite{paeedeh2021improving}, convolutions are also matrix multiplication operations. An $L_{\text{CNN}}$ layers convolutional neural network (CNN) requires $O(L_{\text{MLP}}d_{\text{CNN}}^2)$ operations, where $d_{\text{CNN}}$ is the dimension for the resultant matrix multiplication. Finally, the sequence pooling has two matrix multiplications. Each operation requires $O(nd)$ operations, hence it requires $O(L_{\text{CCT}}nd)$ operations for $L_{\text{CCT}}$ layers.\par
Overall, the time complexity of a CCT model with $L_{\text{CCT}}$ layers is $O(L_{\text{CCT}}nd^2) + O(L_{\text{CCT}}L_{\text{MLP}}d_{\text{MLP}}^2)$.

\section{Experiments}
This section details our numerical study in the BSCD-FSL benchmark \cite{Guo2019ANB} as well as our ablation study to analyze the effect of each learning module. 

\subsection{Datasets}
The BSCD-FSL benchmark \cite{Guo2019ANB} comprises four target datasets: CropDisease \cite{Mohanty2016UsingDL}, EuroSAT \cite{Helber2017EuroSATAN}, ISIC \cite{Codella2019SkinLA} and ChestX \cite{Wang2017ChestXRay8HC} while the base datasets make use of the Mini-ImageNet dataset \cite{Vinyals2016MatchingNF} and the Tiered-ImageNet dataset \cite{Ren2018MetaLearningFS}. We follow the same setting as \cite{Phoo2021SelftrainingFF, Islam2021DynamicDN} where $U$ is constructed from $20\%$ samples of the new domain datasets while the rest are reserved for evaluation. We evaluate both the case of 5-way 1-shot and 5-way 5-shot configurations where 600 runs are performed. Top-1 accuracy and 95\% confidence interval across 600 runs are reported.  

\subsection{Baseline Algorithms}
 ADAPTER is compared with nine prior arts: ProtoNet \cite{Snell2017PrototypicalNF}, MatchingNet \cite{Vinyals2016MatchingNF}, SimCLR \cite{Chen2020ASF}, STARTUP \cite{Phoo2021SelftrainingFF}, Dynamic Distillation Networks \cite{Islam2021DynamicDN}, \textcolor{black}{KT \cite{Li2023KnowledgeTF}, Dara \cite{Zhao2023DualAR}, StyleAdv-FT \cite{Fu2023StyleAdvMS}}. In addition, the numerical results of the transfer learning method as well as the transfer learning method + SimCLR are reported. Numerical results of other methods are obtained from \cite{Islam2021DynamicDN} while we also offer the STARTUP numerical results executed under our computer. ADAPTER's code is developed from the STARTUP environments to ensure fair comparisons and made publicly available for convenient further studies.

\subsection{Implementation details}
We utilize a CCT-7/7x2 backbone with quadruple blocks for all experiments. We add two MLPs with three hidden layers as projection heads for DINO that receive the base-dominant features for the base domain and target-dominant features for the target domain. The Mini-ImageNet dataset is deployed for the base domain, and each dataset in the BSCD-FSL benchmark is deployed for the target domain. The batch size is set to 16. We use the AdamW optimizer to conform to the original implementation of DINO. The weight decay is set to 1e-5, and the weight decay end is set to 0.4. We set the number of epochs to 500 but stopped it at 200 epochs; hence, we could continue an experiment with the same setting for the scheduler without a jump in the learning rate value. For the tiered-ImageNet, we stopped the experiments at 80 epochs.\par

In the final phase, we throw away the projection heads of the DINO. Next, we freeze the backbone network and use a four-layer MLP classifier with ReLU activation functions for the hidden layer on top of the backbone.
Since both inputs of the cross-attentions in the testing phase are from the target domain, the cross-attentions will behave like self-attentions. Therefore, we concatenate the same embeddings for the MLP classifier head. We generalize this idea by replicating the self-attention outputs a few times (four times in our experiments) as the final input of the MLP.
The hidden layer size is four times the number of the input features of the MLP classifier. We utilized the SGD optimizer to train the classifier. The learning rate for fine-tuning is 0.01. Weight decay is 0.001. Momentum and dampening are 0.9 for the SGD. The batch size is 4. The number of episodes is 600. We set the number of epochs to 500 for the 1-shot tests and 100 for the 5-shot test. The self-supervised training phase experiments of Mini-ImageNet are conducted on a GeForce RTX 3060 graphic card with batch size 16, and the Tiered-ImageNet are performed on an RTX A5000 graphic card with batch size 32. All experiments of the few-shot learning phase, including the training of the classifier head, label smoothing, and evaluation, are performed on the RTX A5000 graphic card. Hyper-parameters of ADAPTER are listed in Table \ref{table:hyperparameters}.\par

\begin{table*}[!t]
    \caption{5-way k-shot classification accuracy on $\text{Mini-ImageNet} \rightarrow  \text{BSCD-FSL}$. The * indicates the experiments that we retested on our machine.}
    \label{table:results_mini}
    \centering
    \resizebox{\textwidth}{!}{
        \begin{tabular}{lllllllll}
            \toprule
            Dataset &
            \multicolumn{2}{c}{ChestX} &
            \multicolumn{2}{c}{ISIC} &
            \multicolumn{2}{c}{EuroSAT} &
            \multicolumn{2}{c}{CropDisease} \\
            \cmidrule(lr){2-3}\cmidrule(lr){4-5}\cmidrule(lr){6-7}\cmidrule(lr){8-9}
            Methods & 1-shot& 5-shot    & 1-Shot & 5-Shot & 1-Shot & 5-Shot & 1-Shot & 5-Shot\\
            \bottomrule
            ProtoNet & $21.32 \pm .37$ & $24.72 \pm .43$ & $29.58 \pm .57$ & $42.49 \pm .58$ & $55.32 \pm .88$ & $76.92 \pm .67$ & $52.94 \pm .81$ & $81.84 \pm .68$ \\
            MatchingNet & $20.65 \pm .29$ & $22.62 \pm .36$ & $27.37 \pm .51$ & $33.96 \pm .54$ & $ 54.88 \pm .90$ & $68.00 \pm .68$ & $46.86 \pm .88$ & $63.94 \pm .84$ \\
            Transfer & $22.60 \pm .39$ & $26.51 \pm .43$ & $32.12 \pm .59$ & $43.88 \pm .57$ & $58.14 \pm .83$ & $80.09 \pm .61$ & $ 68.78 \pm .84$ & $89.79 \pm .52$ \\
            SimCLR(Base) & $22.37 \pm .42$ & $26.63 \pm .46$ & $32.43 \pm .56$ & $44.04 \pm .55$ & $ 58.28 \pm .90$ & $80.83 \pm .64$ & $61.58 \pm .88$ & $83.44 \pm .61$ \\
            SimCLR & $23.59 \pm .44$ & $29.56 \pm .49$ & $31.45 \pm .59$ & $42.18 \pm .54$ & $62.63 \pm .87$ & $82.76 \pm .59$ & $69.22 \pm .93$ & $89.31 \pm .53$ \\
            Transfer+SimCLR & $23.72 \pm .44$ & $29.45 \pm .10$ & $31.67 \pm .55$ & $45.97 \pm .54$ & $ 63.91 \pm .83$ & $85.78 \pm .51$ & $70.35 \pm .85$ & $91.10 \pm .49$ \\
            STARTUP & $23.09 \pm .43$ & $26.94 \pm .45$ & $32.66 \pm .60$ & $47.22 \pm .61$ & $63.88 \pm .84$ & $82.29 \pm .60$ & $75.93 \pm .80$ & $93.02 \pm .45$ \\
            STARTUP$^*$ & $23.29 \pm .42$ & $27.11 \pm .47$ & $33.96 \pm .63$ & $47.89 \pm .61$ & $63.56 \pm .85$ & $82.85 \pm .58$ & $75.53 \pm .81$ & $92.90 \pm .44$ \\
            Dynamic Distillation Net. & $23.38 \pm .43$ & $28.31 \pm .43$ & $34.66 \pm .58$ & $49.36 \pm .59$ & $73.14 \pm .84$ & $89.07 \pm .47$ & $82.14 \pm .78$ & $95.54 \pm .38$ \\
            \textcolor{black}{StyleAdv-FT} & $22.64 \pm 0.35$ & $26.24 \pm 0.35$ & $35.76 \pm 0.52$ & $53.05 \pm 0.54$ & $72.92 \pm 0.75$ & $91.64 \pm 0.43$ & $80.69 \pm 0.28$ & $96.51 \pm 0.28$ \\
            \textcolor{black}{KT} & $22.68 \pm 0.60$ & $26.79 \pm 0.61$ & $34.06 \pm 0.77$ & $46.37 \pm 0.77$ & $66.43 \pm 0.93$ & $82.53 \pm 0.66$ & $73.10 \pm 0.87$ & $89.53 \pm 0.58$ \\
            \textcolor{black}{Dara} & $22.92 \pm 0.40$ & $27.54 \pm 0.42$ & $36.42 \pm 0.64$ & $56.28 \pm 0.66$ & $67.42 \pm 0.80$ & $85.84 \pm 0.54$ & $80.74 \pm 0.76$ & $95.32 \pm 0.34$ \\
            ADAPTER & $\boldsymbol{25.28 \pm 0.45}$ & $\boldsymbol{31.00 \pm 0.48}$ & $\boldsymbol{40.66 \pm 0.70}$ & ${55.67 \pm 0.62}$ & $\boldsymbol{73.20 \pm 0.67}$ & $\boldsymbol{92.04 \pm 0.34}$ & $\boldsymbol{85.23 \pm 0.70}$ & $\boldsymbol{96.61 \pm 0.32}$ \\
            \hline
            
        \end{tabular}
    }
\end{table*}
	
\begin{table*}[!t]
    \caption{5-way k-shot classification accuracy on $\text{Tiered-ImageNet} \rightarrow  \text{BSCD-FSL}$. The * indicates the experiments that we retested on our machine.}
    \label{table:results_tiered}
    \centering
    \resizebox{\textwidth}{!}{
        \begin{tabular}{lllllllll}
            \toprule
            Dataset &
            \multicolumn{2}{c}{ChestX} &
            \multicolumn{2}{c}{ISIC} &
            \multicolumn{2}{c}{EuroSAT} &
            \multicolumn{2}{c}{CropDisease} \\
            \cmidrule(lr){2-3}\cmidrule(lr){4-5}\cmidrule(lr){6-7}\cmidrule(lr){8-9}
            Methods & 1-shot& 5-shot    & 1-Shot & 5-Shot & 1-Shot & 5-Shot & 1-Shot & 5-Shot\\
            \bottomrule
            Transfer & $22.46 \pm .41$ & $26.33 \pm .45$ & $29.76 \pm .55$ & $41.27 \pm .58$ & $58.07 \pm .86$ & $81.34 \pm .53$ & $69.94 \pm .87$ & $90.12 \pm .49$ \\
            SimCLR(Base) & $22.28 \pm .40$ & $25.96 \pm .44$ & $31.03 \pm .55$ & $42.91 \pm .55$ & $62.14 \pm .89$ & $81.85 \pm .59$ & $62.45 \pm .90$ & $84.11 \pm .60$ \\
            Transfer+SimCLR & $23.81 \pm .46$ & $30.26 \pm .50$ & $31.71 \pm .55$ & $45.08 \pm .56$ & $58.08 \pm .83$ & $86.08 \pm .47$ & $71.25 \pm .89$ & $91.31 \pm .49$ \\
            STARTUP & $22.10 \pm .40$ & $26.03 \pm .44$ & $29.73 \pm .51$ & $43.55 \pm .56$ & $64.32 \pm .87$ & $85.19 \pm .50$ & $70.09 \pm .86$ & $90.81 \pm .49$ \\
            Dynamic Distillation Net. & $22.70 \pm .42$ & $27.67 \pm .46$ & $33.87 \pm .56$ & $47.21 \pm .56$ & $72.15 \pm .75$ & $89.44 \pm .42$ & $84.41 \pm .75$ & $95.90 \pm .34$ \\
            ADAPTER & $\boldsymbol{25.88 \pm .49}$ & $\boldsymbol{31.85 \pm .49}$ & $\boldsymbol{40.57 \pm .69}$ & $\boldsymbol{55.65 \pm .61}$ & $70.63 \pm .66$ & $\boldsymbol{91.43 \pm .35}$ & $83.80 \pm .72$ & $\boldsymbol{96.55 \pm .34}$ \\
            \hline
        \end{tabular}
    }
\end{table*}
	
	\subsection{Numerical Results}
	Table \ref{table:results_mini} reports our numerical results using the Mini-ImageNet as the base task. It is perceived that ADAPTER produces the most improved results with $1-9\%$ gaps compared to other algorithms across almost all cases. ADAPTER performs equally well for both 1-shot and 5-shot configurations, where the highest margin to its counterparts is observed in the ISIC problem. Note that our experiments are executed under the STARTUP environments to ensure fair comparisons. Table \ref{table:results_tiered} reports our numerical results using the tiered-ImageNet as the base task. The same finding is drawn where ADAPTER outperforms its counterparts with noticeable margins, i.e., $3-4\%$ for the ChextX problem and $6-9\%$ for the ISIC problem. This pattern applies to both the one-shot and five-shot configurations. On the other hand, ADAPTER beats other algorithms in the EuroSAT problem and in the CropDisease problem under the 5-shot setting. ADAPTER is slightly behind the dynamic distillation in the EuroSAT and CropDisease problem under the 1-shot scenario.
	
	\subsection{Ablation Studies}
	An ablation study is carried out to analyze the contribution of each learning module to the overall performance of ADAPTER. It is performed across all datasets, making use of the Mini-ImageNet as the base dataset.\par
    Table \ref{table:ablation_studies_configs} shows which components are used for a specific variation. The accuracies for each target dataset in 1-shot and 5-shot settings are displayed for each variation in Table \ref{table:results_ablation_studies}.
	The first variation is to remove the label smoothing mechanism via the label propagation technique and indicated in the second row of Table \ref{table:results_ablation_studies}. The absence of label smoothing mechanism decreases the performance of ADAPTER significantly by around $0.5\%-1.2\%$. This finding confirms the label smoothing mechanism to overcome the biased responses of the classifiers due to very few samples in $S$.\par
	The second configuration is to include $S$ in the label smoothing mechanism in addition to $Q$. It is shown in the third row of Table \ref{table:results_ablation_studies}. Note that $S$ is removed afterwards and excluded from any accuracy calculations. This modification does not improve the results significantly and often worsens the results. Besides, this configuration also imposes extra computations to induce the adjacency matrix $A$.\par
	Another variation of ADAPTER is attempted where it is trained on the base domain in a supervised manner rather than the self-supervised manner without any cross-attention mechanisms. An MLP classifier, after that, is trained on the support set. This configuration significantly compromises the generalization of ADAPTER and aligns with the finding of \cite{Lu2022SelfSupervisionCB} where the supervised learning step results in the overfitting problem of the target classes. It hinders ADAPTER from generalizing beyond its base classes. This finding also confirms the advantage of bidirectional transformer architecture to dampen the gap between the base task and the target task. \par
	In the last experiment, after training the backbone on Mini-ImageNet, the CCT backbone is fine-tuned simultaneously on Mini-ImageNet, and each target dataset with two distinct linear classifiers to exploit the bidirectional cross-attention mechanism. For evaluation, we feed the target features from self-attention to the classifier as both base and target features. Although the cross-attention improves the predictions in most cases, the few samples in $S$ are not enough to train many parameters of the whole model in a few cases.

\begin{table*}[!t]
    \caption{Settings in ablation studies.  QB, SSRL, and FSL stand for quadruple block, self-supervised representation learning, and few-shot learning, respectively.}
    \label{table:ablation_studies_configs}
    \centering
    \resizebox{\textwidth}{!}{
        \begin{tabular}{llllll}
            Component & Normal & w/o label prop. & label prop w. sup. set & supervised & supervised, fine-tuned \\
            \hline
            \hline
            QB for SSRL & \checkmark & \checkmark & \checkmark & \_ & \checkmark \\
            QB for FSL & \_ & \_ & \_ & \_ & \checkmark \\
            Label smoothing (in FSL) & \checkmark & \_ & \checkmark & \checkmark & \checkmark \\
            S in label smoothing & \_ & \_ & \checkmark & \_ & \_ \\
            Frozen backbone in FSL & \checkmark & \checkmark & \checkmark & \checkmark & \_ \\
            \hline
        \end{tabular}
    }
\end{table*}
	
\begin{table*}[!t]
    \caption{Ablation studies on different configurations for 5-way k-shot classification accuracy on $\text{Mini-ImageNet} \rightarrow  \text{BSCD-FSL}$.}
    \label{table:results_ablation_studies}
    \centering
    \resizebox{\textwidth}{!}{
        \begin{tabular}{lllllllll}
            \toprule
            Dataset &
            \multicolumn{2}{c}{ChestX} &
            \multicolumn{2}{c}{ISIC} &
            \multicolumn{2}{c}{EuroSAT} &
            \multicolumn{2}{c}{CropDisease} \\
            \cmidrule(lr){2-3}\cmidrule(lr){4-5}\cmidrule(lr){6-7}\cmidrule(lr){8-9}
            Methods & 1-shot& 5-shot    & 1-Shot & 5-Shot & 1-Shot & 5-Shot & 1-Shot & 5-Shot\\
            \bottomrule
            ADAPTER & $25.28 \pm .45$ & $31.00 \pm .48$ & $40.66 \pm .70$ & $55.67 \pm .62$ & $73.20 \pm .67$ & $92.04 \pm .34$ & $85.23 \pm .70$ & $96.61 \pm .32$ \\

            ADAPTER (w/o label prop.) & $25.17 \pm .47$ & $30.53 \pm .50$ & $39.67 \pm .67$ & $55.14 \pm .64$ & $71.34 \pm .69$ & $91.10 \pm .35$ & $83.00 \pm .78$ & $95.50 \pm .38$ \\

            ADAPTER (label prop w. sup. set))  & $25.10 \pm .46$ & $30.65 \pm .49$ & $41.06 \pm .68$ & $55.73 \pm .62$ & $72.86 \pm .74$ & $91.56 \pm .36$ & $84.01 \pm .79$ & $96.41 \pm .32$\\

            ADAPTER (supervised) & $21.33 \pm .39$ & $22.78 \pm .42$ & $35.23 \pm .68$ & $47.47 \pm .67$ & $57.86 \pm .85$ & $73.37 \pm .65$ & $66.45 \pm .93$ & $83.34 \pm .60$ \\

            ADAPTER (supervised, fine-tuned)  & $21.73 \pm .39$ & $22.99 \pm .39$ & $34.82 \pm .68$ & $47.81 \pm .66$ & $55.24 \pm .81$ & $70.49 \pm .71$ & $65.04 \pm .88$ & $84.00 \pm .61$ \\	
            \hline
        \end{tabular}
    }
\end{table*}

\subsection{t-SNE analysis}
The t-SNE plots of the features of the CCT backbone for the EuroSAT and CropDisease are shown in Figure 2. The graphs show that the CCT backbone, simultaneously trained on both domains with bidirectional cross-attention, can successfully learn transferable and discriminative features of both domains. While the vanilla transformer is weak in dealing with low-resolution input images, the combination of convolutions and attention mechanisms does not exhibit such a disadvantage. In the second graph, for CropDisease, two classes have significant overlap. Moreover, the model sometimes maps a few samples from one cluster to another cluster. It explains another reason for the effectiveness of label smoothing that assigns the same label to a few labels in the vicinity of each other.

\begin{figure}[h!]
    \centering \label{fig:t-sne}
    \subfloat[EuroSAT]{
        \fbox{\includegraphics[width=0.47\textwidth]{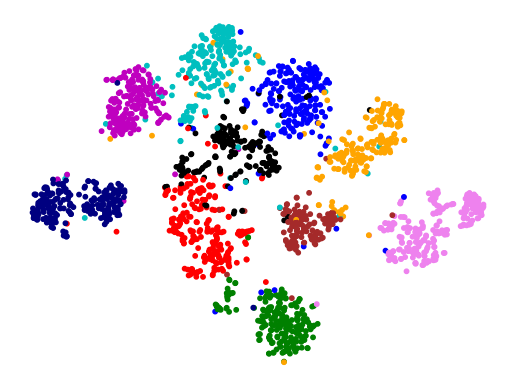}}
    } 
    \quad 
    \subfloat[CropDisease]{
        \fbox{\includegraphics[width=0.47\textwidth]{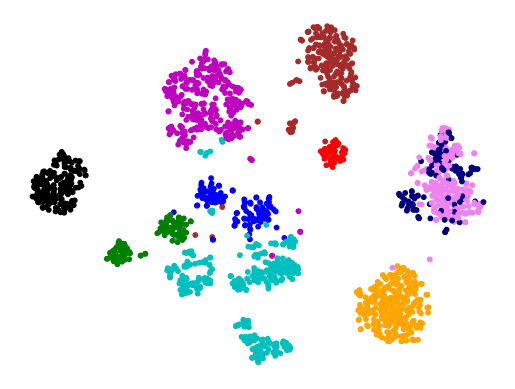}}
    } 
    \caption{t-SNE plots of 10 classes from the test sets of EuroSAT and CropDisease datasets for 2000 samples.} 
\end{figure}
	
	\section{Conclusion}
	This paper proposes a simple but effective solution for cross-domain few-shot learning problems, adaptive transformer network (ADAPTER) constructed under the bidirectional cross-attention transformer architecture. ADAPTER is trained in a self-supervised manner utilizing the bidirectional cross-attention block aligning the labelled base task and the unlabelled target task. ADAPTER applies the label smoothing phase via the label propagation concept to further refine the model's predictions. Our experiments in the BSCD-FSL benchmark have demonstrated the advantage of ADAPTER, where it outperforms prior arts with $1-9\%$ margins in 14 out of 16 experiments across all target datasets in both 1-shot and 5-shot configurations. This finding is consistent regardless of the base dataset, Mini-ImageNet and Tiered-ImageNet.\par
    Our future study is devoted to explore continual cross-domain few-shot learning where a model should be robust against the catastrophic forgetting problem while performing seamless knowledge transfer across the base and the target dataset.

 \begin{table*}[!t]
    \caption{The list of hyperparameters that are used in different phases of the experiments. SSRL, MI, TI, FSL, FT, and s+c stand for self-supervised representation learning, Mini-ImageNet, Tiered-ImageNet, few-shot learning, fine-tuning, and self-attention + cross-attention, respectively.}
    \label{table:hyperparameters}
    \centering
    \resizebox{\textwidth}{!}{
        \begin{tabular}{llllllll}
            Phase & SSRL MI+BSCD-FSL & SSRL TI+BSCD-FSL & Supervised-MI & FSL 1-shot & FSL-5-shot & FT 1-shot & FT 5-shot \\
            \hline
            \hline
            Batch size (MI+BSCD-FSL) & 16 & 32 & 200 & 4 & 4 & 5 & 5 \\
            N. episodes & \_ & \_ & \_ & 600 & 600 & 600 & 600 \\
            N. epochs & 200 & \color{black} 50 & 400 & 500 & 100 & \_ & \_ \\
            N. iterations & \_ & \_ & \_ & \_ & \_ & 500 & 250 \\
            N. epochs in Cosine scheduler & 500 & 500 & \_ & \_ & \_ & \_ & \_ \\
            Optimizer & AdamW & AdamW & SGD & SGD & SGD & SGD & SGD \\
            Learning rate & $2.5 \times 10^{-4}$ & $2.5 \times 10^{-4}$ & 0.025 & 0.01 & 0.01 & 0.01 & 0.01 \\
            Min LR & $5 \times 10^{-7}$ & $5 \times 10^{-7}$ & \_ & \_ & \_ & \_ & \_ \\
            Momentum & 0.9 & 0.9 & 0.9 & 0.9 & 0.9 & 0.9 & 0.9 \\
            Momentum for teacher & 0.996 & 0.996 & \_ & \_ & \_ & \_ & \_ \\
            Dampening & \_& \_ & 0 & 0.9 & 0.9 & 0.9 & 0.9 \\
            Weight decay & $1 \times 10^{-5}$ & $1 \times 10^{-5}$ & $1 \times 10^{-5}$ & 0.001 & 0.001 & 0.001 & 0.001 \\
            Weight decay (cosine scheduler) & 0.4 & 0.4 & \_ & \_ & \_ & \_ & \_ \\
            Warmup epochs & 10 & 10 & \_ & \_ & \_ & \_ & \_ \\
            N. hidden layers (MLP) & 3 & 3 & 0 (linear) & 3 & 3 & 0 & 0 \\
            Input dim. of the head & $2 \times 384$ (s+c) & $2 \times 384$ (s+c) & 384 (self) & $4 \times 384$ (self) & $4 \times 384$ (self)& $2 \times 384$ (s+c) & $2 \times 384$ (s+c) \\
            N. hidden neurons & 2048 & 2048 & \_ & 6144 & 6144 & \_ & \_ \\
            N. local crops & 8 & 8 & \_ & \_ & \_ & \_ & \_ \\
            Output dim. head & 65536 & 65536 & 64 & 5 & 5 & 5 & 5 \\
            Frozen backbone in FSL & \_ & \_ & \_ & \checkmark & \checkmark & \_ & \_ \\
            $\alpha$ (LP) & \_ & \_ & \_ & 0.99 & 0.99 & 0.99 & 0.99 \\
            $\sigma$ (LP) & \_ & \_ & \_ & 50 & 50 & 50 & 50 \\
            rcond for pinv. & & & & 0.1 & 0.1 & 0.1 & 0.1 \\
            N. episodes (FSL) & \_ & \_ & \_ & 600 & 600 & 600 & 600 \\
            \hline
        \end{tabular}
    }
\end{table*}

\bibliography{myreference}
\end{document}